# Multi-label Methods for Prediction with Sequential Data

Jesse Read[a,b,c,1], Luca Martino[d,e], Jaakko Hollmén[c]

[a]*Laboratory of Informatics, École Polytechnique, France*
[b]*Computer Science and Networks Dept. Télécom ParisTech, Université Paris-Sarclay, France*
[c]*Department of Computer Science, Aalto University and HIIT, Helsinki, Finland.*
[d]*Institute of Mathematical Sciences and Computing, São Carlos, Brazil.*
[e]*Image and Signal Processing Group. Universitat de València. Spain.*

**Abstract**

The number of methods available for classification of multi-label data has increased rapidly over recent years, yet relatively few links have been made with the related task of classification of sequential data. If labels indices are considered as time indices, the problems can often be seen as equivalent. In this paper we detect and elaborate on connections between multi-label methods and Markovian models, and study the suitability of multi-label methods for prediction in sequential data. From this study we draw upon the most suitable techniques from the area and develop two novel competitive approaches which can be applied to either kind of data. We carry out an empirical evaluation investigating performance on real-world sequential-prediction tasks: electricity demand, and route prediction. As well as showing that several popular multi-label algorithms are in fact easily applicable to sequencing tasks, our novel approaches, which benefit from a unified view of these areas, prove very competitive against established methods.

## 1. Introduction

Multi-label classification is the supervised learning problem where an instance is associated with multiple class variables (i.e., *labels*), rather than with a single class, as in traditional classification problems. See [1] for a review. The typical argument is that, since these labels are often strongly correlated, modeling the dependencies between them allows methods to obtain higher performance than if labels were modelled independently – at the expense of an increased computational cost.

---

[1]Corresponding author, `jesse.read@polytechnique.edu`

The case of binary labels is most common, where a positive class value denotes the relevance of the label (and the negative or null class denotes irrelevance). Typical examples of binary multi-label classification involve categorizing text documents and images, which can be assigned any subset of a particular label set. For example, an image can be associated with both labels `beach` and `sunset`. This is usually represented in vector form, such that, given a set of labels[2] $\mathcal{L} = \{$`beach`, `urban`, `foliage`, `sunset`, `mountains`, `field`$\}$, then an associated label vector is

$$\mathbf{y} = [y_1, y_2, y_3, y_4, y_5, y_6] = [1, 0, 0, 1, 0, 0]$$

which indicates that the first and fourth labels (`beach` and `sunset`) are relevant. The image itself can be represented by feature vector $\mathbf{x} = [x_1, \ldots, x_D]$, and thus the pair $\mathbf{x}, \mathbf{y}$ represents an image and its associated labels. The multi-label classification paradigm has been successfully considered also in many other domains, such as text, video, audio, and bioinformatics – see [1] and references therein for further examples.

Although binary labels (representing relevance and irrelevance) are enough to represent a huge number of practical problems, the generalization where each label can take multiple values – variously called multi-target, multi-output, or multi-dimensional classification – has also been investigated in the literature (see [3, 4, 5, 6]). In this case each $t$-th 'label' ($t = 1, \ldots, T$) can take on up to $L$ values such as a rating $y_t \in \{1, 2, 3, 4, 5\}$ (where $L = 5$), hour of day $y_t \in \{0, \ldots, 23\}$ (where $L = 24$) and so on, rather than the simple relevance/irrelevance case ($L = 2$). In practice many multi-label algorithms can be applied directly to the general multi-output case, and are always applicable indirectly, following from the fact that any binary number can be represented as any decimal number and vice versa. Figure 1 shows the relationship between these paradigms. Throughout this work, we will continue to use the term multi-label classification for the general case.

Sequential data applications deal with a changing *state* over time, for example of an object or scenario at a particular time index. Approaches to modelling in relevant domains are frequently based on some variety of Markov model, of which detailed overviews are given by [7] and [8].

For example, a traveler's movements among waypoints in a city can be modelled as a series of references to these points, where we can consider $y_t$ as indicating the waypoint at time $t$, then an example of a short path among four points under

---

[2] Such as those in the Scene dataset, see [2].

typical notation[3]
$$y_{1:4} = y_1, y_2, y_3, y_4 = 3, 8, 17, 5$$
where the numbers are unique to each node. The difference in real time between each $t$ and $t+1$ depends on the application (it could be seconds, or minutes, for example). The observation (known often *emission*) available at time point $t$ is represented as vector $\mathbf{x}_t$.

These two problems (the one of multi-label and sequential prediction), have until now mostly received attention as different areas of research. However, they can often be seen not just as related problems, but in fact as identical problems, where the terms 'time index', 'state', 'observation', and 'path' can be interchanged with terms like 'label index', 'label', 'input', and 'label vector', respectively.

We were motivated to take a unified view of these two tasks – multi-label classification and sequential prediction – in a framework that allows the natural application of one to the other. This allows us to apply and further develop suitable techniques from multi-label classification to the domain of sequential prediction, in the form of novel methods that overcoming the disadvantages of hidden Markov models and related approaches by allowing the simultaneous prediction of multiple values across time.

In the first contribution of this work, we compare and contrast typical approaches for modelling of multi-label and sequential data, then draw strong connections between these areas (Section 2). We show that many (if not, most) methods are directly applicable from one problem to the other, and that all methods are applicable in some way, usually only with small modifications to the way the data is preprocessed. We analyze and discuss the relative advantages and disadvantages of each method. Furthering this, we provide a unified view (in Section 3) describing a common framework for multi-label and sequential-data algorithms. We look particularly at the applicability of multi-label methods for obtaining competitive performance and necessary scalability characteristics for sequential prediction. In a novel manner we adapt a Markov-based methodology for multi-label data to create a new method (Viterbi Classifier Chains, Section 4), and discuss its suitability in both domains. This leads us to formulate a further novel approach (in Section 5): Sequential Increasingly-sized Chained Labelsets (SICL), which casts a combination of chain-based and set-based approaches to the sequential problem by taking into account the decay of confidence for points relatively further in the future. In Section 6 we compare against a number of competitive multi-label and sequential methods in empirical evaluations on some real-world sequential-data problems. We find that our novel schemes are competitive and scalable. Finally, in Section 7 we discuss the results, summarize our contributions, draw conclusions and mention promising future work in both areas.

---

[3] Although, in typical Markov-model notation, $y$ is often used to denote the observation or emission, rather than the state.

|       | $L = 2$     | $L > 2$           |
|-------|-------------|-------------------|
| $T = 1$ | **binary**  | **multi-class**   |
| $T > 1$ | **multi-label** | **multi-label**[†] |

[†] also known as multi-output, multi-target, multi-dimensional.

Figure 1: Different classification paradigms: $T$ is the number of class *labels* (or target variables), and $L$ is the number of *values* that each label variable can take.

## 2. Connections Between Multi-label and Sequential Classification Problems

In this work we study the supervised classification task, where a series of inputs is mapped to a series of outputs by a model trained on similar labelled examples (i.e., a training set is available). In the sequential task, classification of the future is often specifically referred to as *prediction* (as opposed to the *estimation* of a current state). In the multi-label context, there is no explicit time context, and therefore the term prediction/estimation are used interchangeably for all outputs.

It should be noted that Markov methods are also used frequently in an unsupervised fashion, which is analogous to clustering in non-sequential data. Although this is also a major task, it is not one that we are directly concerned with in this work.

Also, if the state variable is continuous (i.e., $y_t \in \mathbb{R}$), a natural extension of Markov models are the Kalman and particle filters, which is analogous to multi-output regression. We do not specifically address this case, although many of the connections we look at transfer also easily to the scenario of real-valued outputs.

In Table 1 we outline the parallels between the terminology used in research dealing with the areas of sequential and multi-label data. To the best of our knowledge connection has not been documented to such as extent in the literature. We will start with a discussion on models (Section 2.1) for sequential data, and refer back to these models thereafter as we draw connections from multi-label data (Section 2.2).

*2.1. Models for Sequential Data*

Applications of classification in sequential data abound in the real world and this is echoed in a wealth of scientific literature. Applications include speech, handwriting, and gesture recognition, part-of-speech tagging, daily activity and medical monitoring, fraud detection [9], and traveller modes and movements in an urban setting [10, 11, 12].

Table 1: Notation, and comparison of typical terms in the literature dealing with sequential and multi-label data. Note that indexing with $t$ is more typical of the former, whereas $j$, $k$, or $\ell$ are used to index labels. As the target application of this work involves sequential data, we use the $t$ index henceforth throughout. On the other hand, we use $y_t$ to indicate an output label, and $\mathbf{x}_t$ the inputs, as per multi-label convention and in contrast to many uses in sequential-algorithms, particularly Markov models.

| symbol | sequential data | multi-label data |
|---|---|---|
| $t = 1, \ldots, T$ | time index | label index |
| $L$ | number of states | number of values per label |
| $T$ | sequence length | total number of labels |
| $y_t \in \{1, \ldots, L\}$ | state at time $t$ | value of $t$-th label |
| $\mathbf{x}_{1:T} \equiv \mathbf{x}$ | full emissions | input feature vector |
| $\mathbf{x}_t = [x_1, \ldots, x_D]$ | emission at time $t$ | input subset |
| $y_{1:T} = y_1, \ldots, y_T$ $\equiv \mathbf{y} = [y_1, \ldots, y_T]$ | sequence/path | label vector ($\mathbf{y}$) |
| $\{\mathbf{y}^{(i)}\}_{i=1}^N$ | $N$ sequences | label vectors |
| $\{\mathbf{x}^{(i)}, \mathbf{y}^{(i)}\}_{i=1}^N$ | training data | training data / dataset |

A prominent methodology is that of *Markov model*s, both for estimation and prediction, where the sequence of states generates a corresponding sequence of observations. The classification task can be carried out retrospectively or in real time.

Recal the notation outlined in Table 1 where each state $y_t \in \{1, \ldots, L\}$ at time $t$ is a discrete variable taking one of $L$ values. In a *hidden Markov model* (HMM), each state $y_t$ at time $t$ is seen as generating an observation/emission $\mathbf{x}_t$, in addition to the following state $y_{t+1}$, such that

$$p(y_{1:T}, x_{1:T}) = \prod_{t=1}^{T} p(x_t|y_t) p(y_t|y_{t-1}) = \prod_{t=1}^{T} \phi_{x_t|y_t} \cdot \theta_{y_t|y_{t-1}}, \quad (1)$$

where $\phi_{x_t|y_t}$ and $\theta_{y_t|y_{t-1}}$ return the emission and transition probabilities, respectively. This is illustrated as a probabilistic graphical model in Figure 2a.

In discrete data, these probabilities can be learned simply by counting occurrences the training data (one or more sequences associated with their respective true states) and as such, HMMs can be viewed as a sequential variant of Naive Bayes[4]. Indeed, Naive Bayes is recovered from Eq. (1) simply by replac-

---

[4]In the supervised case.

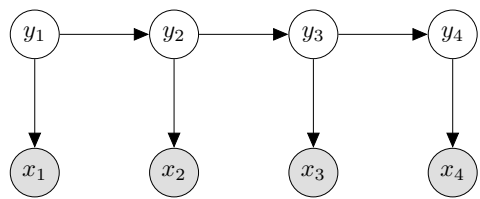
(a) Hidden Markov Model

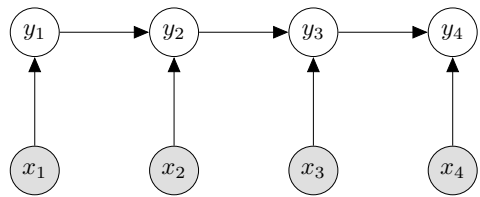
(b) Max. Entropy Markov Model

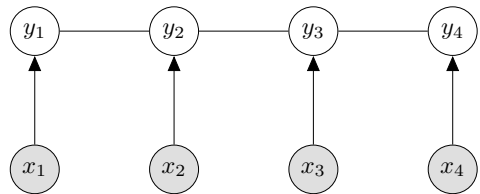
(c) Linear Chain CRF

Figure 2: Markov Models: The HMM is generative, whereas the MEMM (and CRF) are discriminative.

ing the prior $p(y_t|y_{t-1})$ with $p(y_t)$. In the case of a continuous observations ($x_t \in \mathbb{R}$, or $\mathbf{x}_t \in \mathbb{R}^D$), $\phi$ becomes a function, for example a Gaussian, $\phi_{x_t|y_t=k} = \mathcal{N}(x_t|\mu_k, \sigma_k)$, and in this case becomes a sequential variant of centroid, or discriminant analysis classifiers (depending on assumptions about the co-variance matrix).

The task of assigning states to the full sequence out outputs, given the full sequence of inputs, is typically found with the Viterbi algorithm [13],

$$\hat{y}_{1:T} = \underset{y_{1:T}}{\operatorname{argmax}} \, p(y_{1:T}|\mathbf{x}_{1:T}). \tag{2}$$

Approaching the same task from a discriminative point of view, the marginals $p(y_t|\mathbf{x}_{1:t})$ can be modelled directly, resulting in a so-called *Maximum-entropy Markov Model (MEMMs)* [14], see Figure 2b. Note that the *maximum entropy classifier*, where observations are independent, is another name for logistic regression. Therefore a MEMM can be derived from logistic regression with a Markovian assumption among target variables, and can be phrased as conditioning on the previous label as if it were an additional attribute, thus modeling

$$f(\mathbf{x}, y_{t-1}; \mathbf{w}) \propto p(y_t|y_{t-1}, \mathbf{x})$$

with some function $f$, parameterized by some weight vector $\mathbf{w}$.

Generally a MEMM is considered as only involving a forward pass in isolation. A backward pass is no longer straightforward as in HMMs. MEMMs are thus very efficient, but without a backward smoothing pass, they suffer from the *label-bias problem* [14], where a weak probability dilutes the probability inertia across time. For example, a traveller's location becomes uncertain during a section of a journey. This uncertainty is propagated, complicating decisions in future time steps.

*Linear chain conditional random fields* (CRFs) [7, 14] and CRFs in general are a class of model that overcomes the label-bias problem by generalizing feature functions. In the linear chain CRF,

$$f_t(y_t, y_{t-1}, \mathbf{x}; \mathbf{w}) \propto p(y_t|y_{t-1}, \mathbf{x}), \tag{3}$$

where function $f_t$ is typically parameterised by weight vector $\mathbf{w}$.

This is basically a MEMM but with undirected connections among $y_1, \ldots, y_T$ and, on account of these, a global normalization term $\frac{1}{Z}$,

$$p(\mathbf{y}_{1:T}|\mathbf{x}_{1:T}) = \frac{1}{Z(\mathbf{x}_{1:T})} \prod_{t=2}^{T} f_t(y_{t-1}, y_t, \mathbf{x}),$$

whereby the label bias problem is overcome, at the cost of complicating inference. It does, however, have a concave likelihood function that can be learned with e.g., message passing, as the graph is sufficiently unconnected. It is the global normalizer $\frac{1}{Z(\mathbf{x}_{1:T})}$ that makes inference more costly but also precisely the method that overcomes the label bias problem of MEMMs.

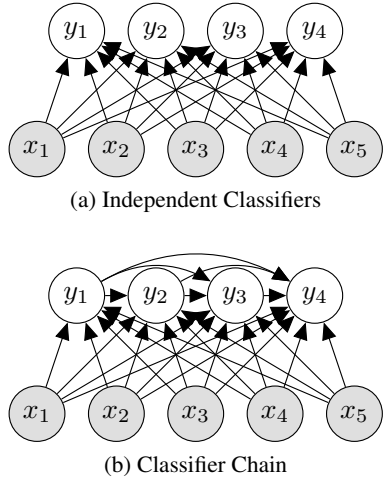

(a) Independent Classifiers

(b) Classifier Chain

Figure 3: Multi-label relevance classifiers – standard (Figure 3a, with independent outputs) and classifier chains (Figure 3b).

*2.2. Models for Multi-label Data, and Connections to Sequential Models*

Multi-label classification is the same setup as discrete time-series classification, where the goal is typically posed in terms of a MAP estimation (for example, [15, 6, 3]),

$$\hat{\mathbf{y}} = \operatorname*{argmax}_{\mathbf{y}} p(\mathbf{y}|\mathbf{x}). \tag{4}$$

This is, in fact, equivalent to Eq. (2). Likewise, a training dataset $\{\mathbf{x}^{(i)}, \mathbf{y}^{(i)}\}_{i=1}^{N}$ consists of instances $\mathbf{x}^{(i)} = [x_1, \ldots, x_D]$ associated with an output vector $\mathbf{y}^{(i)} = [y_1, \ldots, y_T]$. In the multi-label literature labels are typically indexed by $k$, $j$, or $\ell$ (out of a total of $K$ or $L$). However, by instead using $t$ (and $T$, respectively) we can immediately see strong parallels with sequential data.

In the following we consider two major families of multi-label algorithms: using relevance classifiers (commonly called *binary relevance* in the literature where $L = 2$), and *label powerset* methods. These methodologies encompass a great many specific multi-label algorithms.

*2.2.1. Relevance Classifiers*

Using independent classifiers in a series of separate decisions is the natural extension to the single label problem. This approach, often called *binary relevance* in the multi-label literature for case of binary labels, is illustrated in Figure 3a. If understood probabilistically, it factorizes Eq. (4) as

$$\hat{\mathbf{y}} = \underset{\mathbf{y}}{\operatorname{argmax}} \prod_{t=1}^{T} p(y_t|\mathbf{x}), \qquad (5)$$

and predictions can be made separately, in the form

$$\hat{y}_t = \underset{y_t}{\operatorname{argmax}}\, p(y_t|\mathbf{x}).$$

This incurs complexity of $T$ binary decisions at inference time. It is analogous to applying a non-sequential method (e.g., naive Bayes, logistic regression) to sequential data; or like applying a localization algorithm to a tracking problem. It is directly applicable, but due to its analogous limitations, it is roundly criticised (although still used as a benchmark) in the multi-label literature: its model of independence is unlikely to correspond to the true underlying model of most labeling schemes in real-world data, as has also been repeatedly shown empirically in the multi-label literature, particularly with regard to the zero-one loss, which corresponds to the MAP maximizer of Eq. (4) (and one typically used to evaluate multi-label methods).

The method of *classifier chains* [16, 17] overcomes the issue by instead cascading predictions along a chain of labels, see Figure 3b. If understood probabilistically, then the joint probability, under the MAP estimate, factorizes as

$$\hat{\mathbf{y}} = \underset{\mathbf{y}}{\operatorname{argmax}} \prod_{t=1}^{T} p(y_t|\mathbf{x}, y_1, \ldots, y_{t-1}),$$

which corresponds to the chain rule in probability theory. Optimal probabilistic inference can be applied (see [17]), which leads to exponential complexity, in searching all of the $L^T$ possible paths. Greedy inference is also an option, such that

$$\hat{y}_t = h_t(\mathbf{x}, \hat{y}_1, \ldots, \hat{y}_{t-1})$$

for a series of binary classifiers $h_t$ (such that $\hat{\mathbf{y}} = [\hat{y}_1, \ldots, \hat{y}_T] = [h_1(\mathbf{x}), \ldots, h_T(\mathbf{x})]$); thus recovering similar complexity to the standard relevance classifier in practice. There are perhaps dozens of varieties (e.g., [3, 15, 6, 17]) which experiment with different trade-offs between modelling completeness and speed.

*2.2.2. Label Powerset Classifiers*

The *label powerset* method can be seen as a direct approximation of the MAP estimate of $p(\mathbf{y}|\mathbf{x})$ (Eq. (4)), where $\mathbf{y}$ is treated as a single variable rather than factorized in parts (Figure 4a). For a given $\mathbf{x}$, the label powerset method predicts

$$\hat{\mathbf{y}} = \underset{\mathbf{y} \in \mathcal{U}}{\operatorname{argmax}}\, p(\mathbf{y}|\mathbf{x}),$$

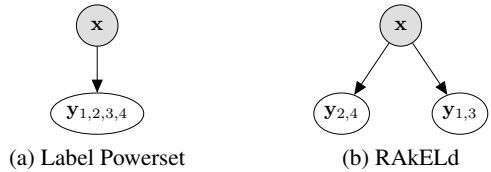

(a) Label Powerset  (b) RAkELd

Figure 4: Multi-label powerset classifiers – standard label powerset (Figure 4a) and an example of RAkELd (Figure 4b).

where $\mathcal{U}$ is a set of *unique* label vectors, typically those from the training data. An obvious major limitation is the large possible size of $|\mathcal{U}| \leq L^T$.

In [18], the RAkEL method splits the set of label indices $1, \ldots, T$ into $M$ subsets of size $k < T$, i.e., $M$ problems,

$$\hat{\mathbf{y}}_m = \operatorname*{argmax}_{\mathbf{y}_m \in \mathcal{U}_m} p(\mathbf{y}|\mathbf{x}),$$

which are combined under a voting scheme. If sets are non-overlapping, this corresponds to simply re-indexing values back into $\hat{\mathbf{y}}$ – which we will denote RAkELd as in [18]. This means that $|\mathcal{U}_m| = k$, limiting complexity to $M \cdot L^k$ ($k$ being some chosen hyper parameter). When the sequence is non-overlapping, there are $M = T/k$ models. This scheme is illustrated in Figure 4b.

It is also possible to simply reduce the size of $\mathcal{U}$ by eliminating rare sequences (i.e., $\mathbf{y} \in \mathcal{U}$ that occurred infrequently in the training data). For example enforcing that $\mathcal{U}$ contain no more than $n$ label vectors (similar to the idea of *support* in itemset mining).

## 3. A Unified View of Multi-label and Sequential-data Methods

The area of multi-label classification has expanded rapidly over recent years and now many techniques have begun overlapping with more established areas like methods for sequential data. Here we develop some of the connections.

The immediately obvious difference between multi-label classification and classification of sequential data lies in the name of the latter: Sequential data is by definition ordered, usually across time[5], whereas for multi-label data, indices are usually considered arbitrarily; it does not matter to the application, for example, if an image is tagged with the label `beach` before or after it is tagged with `urban`.

---

[5]A notable exception is the well-known field of genome sequencing, e.g., [19]

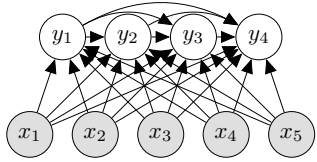

Figure 5: Classifier Chains with $T$ feature functions, compare to Figure 3b and Figure 2b.

Whereas many multi-label methods invest considerable computational resources in uncovering underlying dependency, sequential methods usually simply assume time-based dependency. Sometimes multi-label problems assume some hierarchical dependency, i.e., a hierarchy that predefines some dependency among outputs. But this kind of 'predefined' dependency is often ignored by multi-label methods. Similarly, the Markovian assumption of dependence between all $y_t$ and $y_{t-1}$ is unlikely to be the precise underlying relationship among target variables. In reality, events at time $t$ might have more dependency with time $t-7$, for example if $t$ indexes days of the week. Here, the difference with multi-label classification begins to disappear.

Applying independent classifiers is generally uninteresting for learning sequences because it ignores dependency among states in time. In the multi-label literature, this method (typically called binary relevance in the binary case, BR) is criticised precisely for this reason (for not modeling dependence among labels). That said, in multi-label classification, BR continues to hold its ground as a benchmark method. Probably, because with a strong base classifier, dependence can be modelled 'internally'.

A classifier chain (CC) is essentially a cascading (rather than linear-chain) MEMM; compare Figure 2b and Figure 3b. In fact, it can be viewed as a $T$-order MEMM. Vice versa, a MEMM can be seen as a first-order CC. An obvious difference in applicability between the two is that a MEMM being a chain across time rather than across labels, time-based observations $x_{t+1}$ which cannot be used in the prediction of $x_t$ at time $t$, since it is in the future. This difference is reflected in the figures, where clearly CC has access to all input upon prediction of even the first label. However, this is simply a question of application. If, rather than the raw input, we deal with some generic function $f_t([x_1,\ldots,x_T])$ on the input (possibly a subset), we can represent MEMMs and CCs under the same framework; Figure 5.

Like MEMMs, CC suffers the label bias problem, which in the multi-label literature is called *error propagation* [17] although it refers to the same issue: a poor/uncertain classification of outputs early on, leads to greater difficulty in classifying later outputs accurately later in the sequence/chain. In development of probabilistic perspectives for CC, the error propagation problem was overcome by *proba-* *bilistic* classifier chains (PCC, [17]) and approximations (e.g., [6, 20]). In this case, Eq. (5) is tackled directly and probabilistically (rather than greedily as per original CC). Under the idea of feature functions, a strong connection can in fact be made to general CRFs. Namely, by extending the feature functions of linear-chain CRFs (Eq. (3)) to a full cascade, where $f_t(\mathbf{x}, \mathbf{y}) \propto p(y_t|\mathbf{x}, y_1, \ldots, y_{t-1})$ then the posterior conditional distribution of a CRF as $p(\mathbf{y}_{1:T}|\mathbf{x}_{1:T}) = \frac{1}{Z(\mathbf{x}_{1:T})} \prod_t f_t(\mathbf{x}_t, \mathbf{y}_t)$ (with partition function $Z(\mathbf{x})$). A number of modern multi-label methods have been directly or indirectly based on this methodology. For example [20] presents a fully connected (rather than cascaded) version where feature functions $f_t(\mathbf{x}, \mathbf{y}) \propto p(y_t|\mathbf{x}, y_1, \ldots, y_{t-1}, y_{t+1}, \ldots, y_T)$. This method must carries out many iterations of Gibbs sampling from this same conditional distribution,

$$y_t \sim p(y_t|\mathbf{x}_t, y_1, \ldots, y_{t-1}, y_{t+1}, \ldots, y_T),$$

to converge; of which the final iterations collect the marginal probabilities for each label. The use of Gibbs sampling reflects the fact that in a densely connected CRF as in [20], message passing is no-longer feasible and different methods are necessary. A number of sampling methods were introduced also for the directed-graph representation (Figure 3b, Figure 5), e.g., [15, 6, 21] also for this purpose: to take samples from the predictive density. [22] gives a CRF interpretation where also feature variations are included in the functions. Other methods use more sparsely-connected structures like the classifier trellis (CT) approach [23], which also experiments with different node orderings at training time. Approximate inference (such as, for example, [15]) can also be used on this structure. When converged, a CRF/PCC method should not suffer from the label bias problem as do MEMMs and greedy-inference CC, although inference is inherently more costly.

The label powerset method (LP) on the other hand, as it does not factorize the probability distribution (unlike P/CC in Eq. (5)), suffers from a number of serious drawbacks preventing ready application to sequential data (and hence, commonly-used parallels in sequential data methodology), namely

1. very high computational complexity, $\mathcal{O}(\min(N, L^T))$ possible unique observed labelsets;
2. it cannot be applied as-is to sequential data in a filtering/real-time prediction context, since the end of the sequence ($y_T$) is needed to begin inference; and
3. only pre-observed training sequences can be predicted.

The first two issues are similar to a fully cascaded Bayes-optimal classifier chain with regard to inference, but since a single classification is carried out, rather than multiple individual classifications, obtaining approximations via sampling like is not possible like in CC methods. In some multi-label document corpora, only a few label vectors occur frequently, but this is not usually the case in sequential applications, and therefore the third issue is already very limiting.

Interestingly, whereas these drawbacks mostly rendered LP out of consideration for sequential data applications, the multi-label community has made a series of advances to make it considerably more tractable. RAkEL (see Section 2.2) is one of these. We note, that the particular variety of RAkEL$d$ (where the $d$ stands for *disjoint*), which splits up the labelset into (disjoint) blocks, could in fact be applied in a sequential manner, if labelsets are kept in sequential time order rather than randomized is in the original algorithm. We return to this idea in Section 5.

There is one group of methods already used for sequential learning related to LP: lazy/prototype-based methods, such as explored in [24]. In instance-based methodology, a new set of observations is compared to a set of existing (prototype) vectors and the closest associated output (or a mixture of the closest) is chosen. In [24] such a method is used for route recognition, using prototype routes and a subsequence matching method. In this particular case, LP-like methods are suitable, since it is already assumed that the set of routes taken from a particular point, is a relatively small subset of possible routes, for which some case already exists in the training data.

*3.1. Dealing with sequences*

Finally, there is another important difference in sequential data learning: the training sequences $\mathbf{x}^{(i)}, \mathbf{y}^{(i)}$ are typically of different lengths. We can speak of length of the $i$-th sequence being $T_i$ which may be difference to other sequences. This arises naturally in sequential data, for example natural-language sentences and paragraphs are not of uniform length, and the length of a traveller's trajectory varies depending on the journey.

To be able to apply multi-label classifiers under this scenario, we recast the problem into blocks. For some window of length $\tau$, we create for the $i$-th sequence new instances $\mathbf{x}_t = (\mathbf{x}_t, y_{t-\tau}, \ldots, y_{t-1})$ associated with labels $\mathbf{y}_t = [y_t, \ldots, y_{t+\tau}]$ for some prediction horizon $\tau$, up until $T_i$. This forms a new dataset of $\sum_{i=1}^{N} \frac{T_i}{\tau}$ instances. In this way, we can any multi-label learner to sequential data, and $\tau$ simply becomes the number of labels ($T$) in the transformed dataset. This transformation method is illustrated in Figure 6.

Various existing multi-label methods (RAkEL is a notable example) already divide the sequence (label vector) into subsets. We develop a related method (see Section 5), which additionally links blocks together, such that information is not separated (as it otherwise would be with, for example, RAkEL). And in fact, most multi-label methods already transform datasets before operating on them (in fact, most of those discussed already are typically called 'problem transformation' methods [25]). Therefore, this block method can simply be seen as another kind of

$$\text{sequence} = \begin{bmatrix} \mathbf{x}^{(1)} & y^{(1)} \\ \mathbf{x}^{(2)} & y^{(2)} \\ \mathbf{x}^{(3)} & y^{(3)} \\ \mathbf{x}^{(4)} & y^{(4)} \\ \mathbf{x}^{(5)} & y^{(5)} \\ \mathbf{x}^{(6)} & y^{(6)} \end{bmatrix} \quad \text{transformation} = \begin{bmatrix} (\mathbf{x}^{(2)}, \mathbf{x}^{(3)}, y^{(1)}, y^{(2)}) & [y^{(3)}, y^{(4)}] \\ (\mathbf{x}^{(3)}, \mathbf{x}^{(4)}, y^{(2)}, y^{(3)}) & [y^{(4)}, y^{(5)}] \\ (\mathbf{x}^{(4)}, \mathbf{x}^{(5)}, y^{(3)}, y^{(4)}) & [y^{(5)}, y^{(6)}] \end{bmatrix}$$

Figure 6: Transformation of a sequence of length $T_i = 6$ (left) to a multi-label dataset, with time horizon $\tau = 2$ (and thus $T = 2$, right). Each row of the transformed dataset is an instance consisting of vector pairs $\mathbf{x}_t, \mathbf{y}_t$. In words, a past time horizon is associated with a future time horizon, without violating the online constraint that future emissions cannot be used as input.

transformation. Note that in the case where $(N \mod \tau) \neq 0$, we can simply pad the last block with extra labels. This depends on the data, we discuss the specifics in Section 6.2. For the case of a traveller moving among nodes, for example, padding with the final node is a sensible choice.

*3.2. Evaluation Metrics*

Typically in multi-label experiments, the metrics *subset $0/1$ loss* and *Hamming loss* are among the most commonly used metrics in the multi-label literature (e.g., [18, 20]). The $0/1$ loss,

$$\ell_{\mathbf{y},\hat{\mathbf{y}}}^{0/1} = 1_{[\hat{\mathbf{y}} \neq \mathbf{y}]}$$

(commonly known in its payoff form as *exact match*) is often preferred in multi-label classification because it rewards methods that invest in regularization or explicit modelling of label dependencies (such methods are a typical focus). However, it since a single bit of difference leads to $0$ for a particular example, it is too hash for most sequence-labelling tasks, where perfect labelling is not expected. Hamming loss – also used frequently in multi-label evaluations – evaluates each label separately, such that for a true and predicted labelset vector,

$$\ell_{\mathbf{y},\hat{\mathbf{y}}}^{\mathsf{HL}} = \frac{1}{T} \sum_{t=1}^{T} 1_{[\hat{y}_t \neq y_t]}.$$

Essentially, it encourages maximizing the marginal probabilities in isolation. There exist in fact many other metrics used in binary multi-label classification (e.g., Jaccard index, micro and macro F-measures), but these to not transfer easily to the multi-output case where labels are not necessarily binary.

Neither Hamming loss nor Exact match are particularly suitable for sequential evaluation, since they ignore proximal dependence. For example, if $\mathbf{y} =$ $[0, 8, 2, 9, 7]$ and $\hat{\mathbf{y}} = [8, 2, 9, 7, 0]$, then Hamming loss is 1 (indicating a maximally incorrect prediction), even though most of the sequence is correct, only out of alignment by one timestep. More typical of, and suitable for, time series are *edit distances*, for example the *Levenshtein distance* (e.g., in [26]), which allows for insertion and deletion as well as substitution. The recursive formula is

$$\ell_{\mathbf{y},\hat{\mathbf{y}}} = \begin{cases} \max(i,j) & \text{if } \min(i,j) = 0 \\ \min \begin{cases} \ell_{\mathbf{y},\hat{\mathbf{y}}}(i-1, j) + 1 \\ \ell_{\mathbf{y},\hat{\mathbf{y}}}(i, j-1) + 1 \\ \ell_{\mathbf{y},\hat{\mathbf{y}}}(i-1, j-1) + 1_{y_i \neq y_j} \end{cases} & \text{otherwise,} \end{cases}$$

and gives a distance of $\ell_{[0,8,2,9,7],[8,2,9,7,0]} = 2$ (one insertion and one deletion).

Note that Hamming loss is fact the average Hamming *distance*, which is a simple edit distance that allows only for substitution. In contrast, the *longest common subsequence* metric allows only insertion and deletion, but not substitution.

## 4. Viterbi Classifier Chains

Various methods have been applied as approximate inference on highly connected models [27, 6, 21]. Moreover, a number of search variants for optimal and approximate inference have been proposed for classifier chains, including: [27]'s $\epsilon$-approximate inference, based on performing a depth-first search in the probabilistic tree with a cutting-off list; 'beam search' [21], a heuristic search algorithm that speeds up inference considerably; and Monte Carlo search [6]. The optimal and exhaustive inference procedure has been considered in [17].

Instead, in this section we look at an approach that approximates the chain structure, but still allows the Bayes-optimal inference: Viterbi Classifier Chains (VCC). This simpler graphical model is shown in Figure 7. VCC still takes in account the label dependence but allows the Bayes-optimal inference in a faster and more scalable way than [17]. More specifically, given a specific sequence $y_1, \ldots, y_T$, in this case we consider the simplified factorization, like a MEMM, such that

$$p(\mathbf{y}|\mathbf{x}) = p(y_1|\mathbf{x}) \prod_{t=2}^{T} p(y_t|\mathbf{x}, y_{t-1}), \quad (6)$$

where the dependence structure is simpler.

Unlike the exponentially branching tree of paths of CC, VCC's inference can be represented entirely as a trellis diagram, Figure 8. Hence, the goal is to find the optimal label vector and maximize Eq. (4) (the MAP estimate). And this is done considering more than just the current time (as in a MEMM). The vector $\hat{\mathbf{y}}$ coincides with the optimal path through the trellis, as exemplified in Figure 8, and it is well-known that this path can be efficiently obtained using the *Viterbi algorithm* [13].

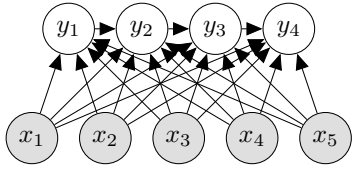

Figure 7: The Viterbi Classifier Chain (VCC)

Regarding differences with Viterbi inference on a Markov chain, there are several important differences. VCC is slightly unusual from the point of view of Markov models, since the state at each 'timestep' (i.e., label) $y_t$ may be of a different shape and form than the previous timestep. Furthermore, we are dealing with the block-transformation point of view that we elaborated in Section 3.1: a dataset of equally-sized blocks, rather than relatively smaller number of variable-length sequences. The concept of a 'base classifier' is rarely considered in sequential-learning literature, but is common in multi-label classification. The same strategy can vary with different base classifiers.

The metric of each branch is defined by $f_t(y_t|\mathbf{x}, y_{t-1})$, an approximation of $p$ provided by any multi-*class* classifier (such as those used individually in BR). Note that the transition probability $p_t(y_t|\mathbf{x}, y_{t-1})$ depends on the index $t$. Also, with regard to sequential data, this scenario is the special case of a generic HMM since we always "observe" the same instance $\mathbf{x}$, for all $t = 1, \ldots, T$, but – as discussed – is only a question of application.

The Viterbi Classifier Chain (VCC) algorithm is summarized in Table 2. We postpone further discussion, and reflections of empirical results until Section 6, but in general we can make the following observation. The connectivity complexity (number of connections) versus inference complexity (e.g., number of iterations) is inherently connected to the *base learner*, a concept predominantly discussed in regard to multi-label classification, since it is not as typical to speak of a 'base classifier' in sequential learning in this sense of applying an 'off-the-shelf' method. In our later analysis, it is clear that a non-linear base learner can counteract the need for extensive connectivity. In other words, investment in connectivity can make up for lack of investment in a base classifier, and vice versa. VCC invests more in inference, whereas the method we present in the following section places more emphasis on connectivity, i.e., modelling labels together.

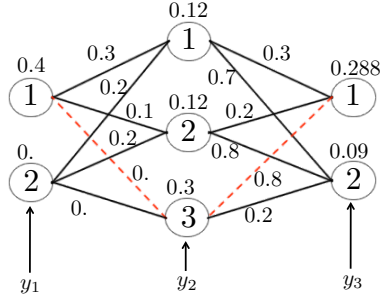

Figure 8: Example of the trellis diagram corresponding to VCC, with 3 class labels $y_t$ ($t = 1, \ldots, 3$), which can take 2, 3, and 2 values, respectively. The best path, $\mathbf{y} = [1, 3, 1]$, with probability 0.288, is shown with dashed red lines.

---

**1. Initialization:**
- Obtain $f_1(y_1|\mathbf{x})$ by a multi-class classifier.
- Set $\delta_1(i) = f_1(y_1 = i|\mathbf{x})$ and $\psi_1(i) = 0$ with $i = 1, \ldots, L_1$.

**2. Recursion:**
For $t = 2, \ldots, T$:
  For $k = 1, \ldots, L_t$:
  - Obtain $f_t(y_t|\mathbf{x}, y_{t-1})$ by a multi-class classifier.
  - Set
  $$\delta_t(k) = \max_{1 \leq i \leq L_k} \delta_{t-1}(i) f_t(y_t = k|\mathbf{x}, y_{t-1} = i),$$
  $$\psi_t(k) = \arg\max_{1 \leq i \leq L_k} \delta_{t-1}(i) f_t(y_t = k|\mathbf{x}, y_{t-1} = i).$$

**3. Output ($\hat{\mathbf{y}}_{VCC} = [\hat{y}_1, \ldots, \hat{y}_t, \ldots, \hat{y}_T]^\top$):**
- $\hat{y}_T = \arg\max_{1 \leq i \leq L_t} \psi_T(i)$
- $\hat{y}_t = \arg\max_{1 \leq i \leq L_t} \psi_t(\hat{y}_{t-1})$

Table 2: Viterbi Classifier Chain (VCC).

## 5. Sequential Increasingly-sized Chained Labelsets

As we explained earlier, multi-label classifiers can be applied easily to the task of sequential prediction. The order of labels itself is not a factor in applicability, weather it be inference, or predicting ahead. Assuming neighbouring (Markov) time dependency has in many cases no particular advantage over arbitrary dependence of equivalent connectedness. However, there is a peculiarity of sequential data with regard to the problem of predicting into the future: generally accuracy will decrease the further we predict into the future.

Figure 10 shows how prediction becomes increasingly difficult with respect to the time horizon. This is entirely expected and fully intuitive. However, we notice in particular that methods based on individual models (independent classifiers (IC), MEMM, CC) are relatively better at predicting the immediate future, whereas the methods based on modelling labels together is relatively better at predicting further into the future. We additionally note that among these methods whose performance is plotted in the figure, only RAkELd treats labels in a different way to their default time order (namely, randomly). Despite this randomness regarding labels, its accuracy nevertheless also decreases with regard to the future horizon.

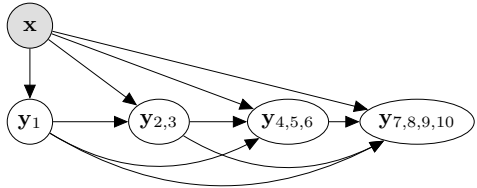

Figure 9: The Sequential Increasing-size Chained Labelsets (SICL) method, illustrated for the case where $T = 10$, $\alpha = 1$, such that the $m$-th set contains $\alpha m = m$ labels.

From this observation we formulated a novel method: Sequential Increasingly-sized Chained Labelsets (SICL), which combines the respective advantages found in different methods. On the one hand, we create subsets of the labelset similarly to RAkELd (described in Section 2.2.2). Unlike RAkELd, however, the original time order of labels is maintained, and we design the sets such that they become increasingly larger nearer the end. In other words, label indices further in the future are grouped into relatively larger subsets. We determine this increment linearly with a hyper-parameter $\alpha$: the $m$-th set contains $\alpha m$ label indices. For example, where $\alpha = 1$, the initial set contains only one label, the second set contains two labels, the third set three labels, and so on. Therefore, classification begins very step-specific with an individual model for the first step, and becomes increasingly prototype-orientated with respect to the prediction horizon. Furthermore, we link these sets together such that information from the first label can be carried along the chain, but the more uncertain predictions (further in the future) are made nearer the end of the chain, thus error propagation is reduced. This method is illustrated in Figure 9. Note that, as with RAkELd, each set is essentially serving as a meta label taking a single value which represents the value of many labels.

In addition to the insight provided by Figure 10, this strategy has additional theoretical backing. In earlier work [28], we described how CC gets its power from leveraging other labels into a higher space (rather than leveraging dependency-information as earlier postulated in the literature). Therefore, labels further down the chain can benefit relatively more. In the sequential context, these are precisely the labels which need additional predictive power. Meanwhile, [15] explains how LP-methods such as RAkEld get their predictive power from modelling la-bels together, at a cost of flexibility (i.e., at a cost to marginal dependence of each label/state). The advantage of our methods is thus twofold: additional predictive power from being inside a meta node, and from high-level features in the form of the earlier nodes in the chain, meanwhile the initial labels are afforded more flexibility.

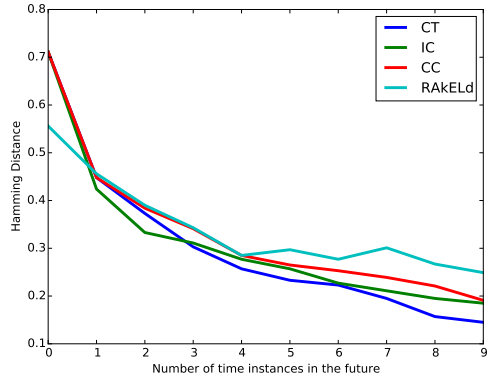

Figure 10: Error plotted against timestep within the prediction window, i.e., further along the horizontal axis represents further into the future. On the Traveller data with $\tau = 10$ prediction window, Naive Bayes as base classifier. IC signifies independent classifiers which is BR in the multi-label context.

## 6. Experiments

We carry out experiments on the task of *multi-step ahead prediction*. In sequential terms, given the current state and observation, we wish to estimate $\tau$ future states. There are several typical strategies for this [29]: In the so-called *iterated strategy*, the state is predicted one step ahead, and this prediction is used as input to estimate the second state, and so on. This is essentially rolling forward an HMM with a flat emission probability. In a *direct strategy*, all future time steps within the time horizon of interest are estimated at once. Binary relevance and label-powerset methods are a direct strategy, whereas classifier chains is a combination of the two approaches, since the input *and* earlier predictions are used in the prediction of future states. A straightforward approach to the iterated strategy is using traditional hidden Markov models and simply 'rolling forward' using only the transition probability and flat emission probability.

*6.1. Datasets*

We use the following data sources (summarized in Table 3):

Table 3: Data sources, with number of timesteps $T$ and states $L$ (see Table 1).

| name | $L$ | $T$ |
|---|---|---|
| Electricity | 2 | 45,312 |
| Traveller 1 | 100 | 122,347 |
| Traveller 2 | 100 | 118,494 |
| Traveller 3 | 100 | 11,717 |

*Electricity.* A relatively well-known time series dataset described by [30]. This data was collected from the Australian New South Wales Electricity Market. In this market, prices are not fixed and are affected by demand and supply of the market. They are set every five minutes. The dataset contains 45,312 instances. The class label identifies the change of the price relative to a moving average of the last 24 hours (UP or DOWN). Available at http://moa.cms.waikato.ac.nz/datasets/. Unlike many other uses of this dataset, such as in data-streams frameworks like MOA, we try to predict labels for *multiple* time steps in the future.

*Traveller.* A set of data streams we prepared from GPS measurements recorded continuously for two weeks by personal smartphones used by participants in a project to gather sensory data. GPS measurements were taken regularly (around every 10 seconds) and averaged by minute. The average position of each minute is snapped to the nearest *node*, which is one of 100 points clustered particularly on an initial batch of 20% of the traveller's data. We found $k$-means clustering to be effective (with $k = 100$). Each node thus represents a geographical point the subject frequently visits or passes through. The current node is associated with the current GPS coordinates (latitude, longitude), the day of the week (1—7), and time of day in hours (0.00—23.98, where the maximum number refers to time 23:59). In total we created three data streams from three different travellers living and working in the greater Helsinki region in Finland. Figure 11 shows a visualization of part of the trace of one of the travellers. The traveller participants are aware of the use of the data in this analysis. Their privacy is retained to the extent that we do not reveal any identifying information regarding the traveller or their clustered nodes.

*6.2. Experiment procedure*

Each of the data sources in Table 3 can be viewed as a continuous stream of pairs $\mathbf{x}_t, y_t$. From this we create multi-label datasets as in Section 3.1, with both $\tau = 5$ and $\tau = 10$. Similar approaches have been used in the literature, for example [31] for modelling vessel trajectory.

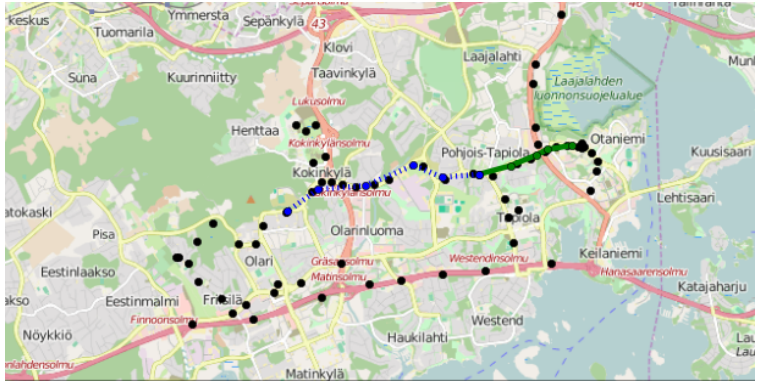

Figure 11: A visualization of part of the Traveller dataset for $\tau = 5$. Black points show the nodes (i.e., the values that $y_t$ can take at each timestep); the green line is the true route over 5 timepoints $y_{t-4}, \ldots, y_t$; the blue dashed line shows links the *predicted* nodes over the following timesteps $\hat{y}_{t+1}, \ldots, \hat{y}_{t+5}$.

For the experiments, we combine existing methods and implement the novel methods in the MEKA framework [32]. MEKA supports a plethora of base classifiers. We carry out experiments both with naive Bayes and decision trees.

Because of the stacked nature of the datasets (where the past and future time horizon of interest is contained within each instance), we carried out a randomized two-fold evaluation, rather than incremental evaluation.

The evaluation metrics we consider have been detailed in Section 3.2.

*6.3. Results*

Table 4 compares the accuracy of a MEMM and CC, both with the nodes under default ordering (by time), and under random order. Clearly, time-order provides no advantage in this context (prediction into the future). Predictive performance is either identical or slightly worse under the original time order. Rather than investing significant effort into inference on a fixed time order, it can pay off to instead invest into finding improvement in label order for the model. This helps justify the general application of multi-label methods (which do not assume time or sequence-based order) to sequential data.

In Table 6, results are shown for a variety of methods already discussed (MEMM, CC, RAkELd, CT, PCC), including benchmark independent classifiers (IC), compared with our novel methods (VCC, SICL). For RAkELd we set $k = 3$ (subset size) as recommended in [18]. We note that PCC is representative of CRFs and a number of modern multi-label methods (such as [17, 20, 6, 22]), and it is basically only the level of approximation varies: we consider a fully-cascaded chain (as in

Table 4: Accuracy (and Exact Match for Traveller). Base classifier `J48`-decision teres. Cross-validation.

| Ordering | Time | | Random | |
| --- | --- | --- | --- | --- |
| Method | MEMM | CC | MEMM | CC |
| Electricity | 0.672 | 0.655 | 0.671 | 0.671 |
| Traveller | 0.144 | 0.150 | 0.153 | 0.155 |

Table 5: Methods used in the experimental comparison, with associated reference (except where presented as a novelty in this paper) and parameters (where applicable).

| Method (Abbr.) | Reference | Parameters |
| --- | --- | --- |
| Independent classifiers (`IC`) | [25] | |
| Classifier Chains (`CC`) | [16] | |
| Maximum Entropy Markov Model (`MEMM`) | [14] | |
| Random $k$-labelled Subsets (`RAkd`) | [18] | $k = 3$ |
| Classifier Trellis (`CT`) | [23] | $\ell = 2$ |
| Probabilistic Classifier Chains (`PCC`) | [17] | $M = 100$ |
| Viterbi Classifier Chains (`VCC`) | | |
| Seq. Increasingly-sized Chained Lablesets (`SICL`) | | $\alpha = 3$ |

[17]), and a maximum of 100 Monte-Carlo samples (as suggested in [6]) to limit complexity. For CT, we consider a trellis of density 2, ordered according to mutual information. For SICL, we set $\alpha = 3$. Other methods do not require additional hyper-parameterization. We have summarized all methods, associated references, and hyper parameters, in Table 5. We remind that all methods are implemented in the MEKA framework [32].

*6.4. Discussion*

With Naive Bayes as a base classifier, our method SICL is the most competitive overall, particularly on the more stricter Levenshtein and Zero/One loss metrics. Under Hamming Loss, it is second only to CT. Since it outperforms RAkELd (which is very similar, except random uniformly-sized blocks) we can conclude that the mechanism we use (of sets of increasing size) is effective, as opposed to random sets. Also, it outperforms CC, which contains the chaining mechanism, but no subsets. We note that RAkELd performs best on the electricity dataset; this dataset has fewer combinations and can be more effectively learned by this method.

With a more powerful base classifier (namely, decision trees), the simpler classifiers with less dependency encoding (e.g., MEMM, VCC) become relatively more

Table 6: Results on dataset-versions of Electricity and Traveller data streams, each stacked under horizon $\tau = 5$ and $\tau = 10$. Performance is measured under 2-fold cross validation and the average is shown under three different metrics. Per-dataset ranks of each method are shown after each result, separated by a space, and the average ranks are shown in the final row. Results are shown in the table only to 3 decimal places, thus in some cases ranks may vary for the same value in the table.

Levenshtein Distance:

| Dataset | IC | CC | MEMM | VCC | RAkd | PCC | CT | SICL |
|---|---|---|---|---|---|---|---|---|
| Elec05 | 0.278 7 | 0.283 8 | 0.271 2 | 0.272 3 | 0.270 1 | 0.277 6 | 0.272 3 | 0.272 3 |
| Elec10 | 0.367 1 | 0.377 5 | 0.376 4 | 0.372 3 | 0.367 1 | 0.379 6 | 0.386 7 | 0.386 7 |
| Tr1-05 | 0.621 7 | 0.600 2 | 0.600 2 | 0.600 2 | 0.660 8 | 0.600 2 | 0.600 2 | 0.580 1 |
| Tr1-10 | 0.661 6 | 0.651 4 | 0.668 8 | 0.661 6 | 0.615 2 | 0.642 3 | 0.651 4 | 0.614 1 |
| Tr2-05 | 0.588 8 | 0.564 6 | 0.535 2 | 0.538 4 | 0.570 7 | 0.557 5 | 0.536 3 | 0.499 1 |
| Tr3-05 | 0.521 7 | 0.499 5 | 0.495 3 | 0.497 4 | 0.588 8 | 0.499 5 | 0.492 2 | 0.478 1 |
| Tr2-10 | 0.696 8 | 0.674 7 | 0.662 5 | 0.658 4 | 0.650 2 | 0.662 5 | 0.657 3 | 0.621 1 |
| Tr2-10 | 0.644 7 | 0.628 5 | 0.600 1 | 0.610 3 | 0.670 8 | 0.618 4 | 0.600 1 | 0.630 6 |
| avg rank | 6.38 | 5.25 | 3.38 | 3.62 | 4.62 | 4.50 | 3.12 | 2.62 |

Hamming Loss:

| Dataset | IC | CC | MEMM | VCC | RAkd | PCC | CT | SICL |
|---|---|---|---|---|---|---|---|---|
| Elec05 | 0.278 7 | 0.283 8 | 0.271 2 | 0.272 3 | 0.270 1 | 0.277 6 | 0.272 3 | 0.272 3 |
| Elec10 | 0.370 1 | 0.377 6 | 0.376 4 | 0.372 3 | 0.370 1 | 0.379 7 | 0.376 4 | 0.386 8 |
| Tr1-05 | 0.626 7 | 0.611 3 | 0.613 6 | 0.612 4 | 0.671 8 | 0.612 4 | 0.600 2 | 0.591 1 |
| Tr1-10 | 0.666 6 | 0.663 5 | 0.675 8 | 0.669 7 | 0.629 1 | 0.652 3 | 0.659 4 | 0.630 2 |
| Tr2-05 | 0.590 8 | 0.566 6 | 0.537 2 | 0.540 4 | 0.576 7 | 0.559 5 | 0.538 3 | 0.500 1 |
| Tr3-05 | 0.524 7 | 0.500 4 | 0.498 3 | 0.500 4 | 0.600 8 | 0.500 4 | 0.495 2 | 0.482 1 |
| Tr2-10 | 0.700 8 | 0.681 7 | 0.666 5 | 0.663 4 | 0.659 2 | 0.669 6 | 0.662 3 | 0.630 1 |
| Tr2-10 | 0.649 7 | 0.635 5 | 0.613 2 | 0.615 3 | 0.685 8 | 0.625 4 | 0.600 1 | 0.644 6 |
| avg rank | 6.38 | 5.50 | 4.00 | 4.00 | 4.50 | 4.88 | 2.75 | 2.88 |

Zero/One Loss:

| Dataset | IC | CC | MEMM | VCC | RAkd | PCC | CT | SICL |
|---|---|---|---|---|---|---|---|---|
| Elec05 | 0.527 8 | 0.512 7 | 0.493 2 | 0.494 3 | 0.500 6 | 0.498 5 | 0.494 3 | 0.491 1 |
| Elec10 | 0.945 7 | 0.943 6 | 0.927 3 | 0.922 1 | 0.930 4 | 0.946 8 | 0.938 5 | 0.924 2 |
| Tr1-05 | 0.971 8 | 0.934 4 | 0.948 6 | 0.942 5 | 0.891 1 | 0.928 3 | 0.950 7 | 0.927 2 |
| Tr1-10 | 0.986 8 | 0.972 4 | 0.977 7 | 0.973 5 | 0.951 1 | 0.965 3 | 0.974 6 | 0.961 2 |
| Tr2-05 | 0.867 8 | 0.843 7 | 0.790 2 | 0.789 1 | 0.827 5 | 0.834 6 | 0.800 3 | 0.800 3 |
| Tr3-05 | 0.787 7 | 0.758 6 | 0.746 2 | 0.748 3 | 0.873 8 | 0.753 5 | 0.748 3 | 0.700 1 |
| Tr2-10 | 0.934 7 | 0.915 6 | 0.900 1 | 0.911 3 | 0.953 8 | 0.900 1 | 0.911 3 | 0.913 5 |
| Tr2-10 | 0.961 6 | 0.921 2 | 0.928 4 | 0.929 5 | 0.978 8 | 0.915 1 | 0.922 3 | 0.962 7 |
| avg rank | 7.38 | 5.25 | 3.38 | 3.25 | 5.12 | 4.00 | 4.12 | 2.88 |

Table 7: Results as in Table 6 but with a decision tree classifier as base classifier.

Levenshtein Distance:

| Dataset | IC | CC | MEMM | VCC | RAkd | PCC | CT | SICL |
|---|---|---|---|---|---|---|---|---|
| Elec05 | 0.300 7 | 0.293 5 | 0.279 1 | 0.279 1 | 0.312 8 | 0.298 6 | 0.288 4 | 0.287 3 |
| Elec10 | 0.345 4 | 0.354 7 | 0.329 1 | 0.329 1 | 0.360 8 | 0.353 6 | 0.340 3 | 0.346 5 |
| Tr1-05 | 0.538 5 | 0.543 7 | 0.525 3 | 0.523 2 | 0.629 8 | 0.539 6 | 0.530 4 | 0.517 1 |
| Tr1-10 | 0.599 7 | 0.589 5 | 0.577 3 | 0.575 2 | 0.619 8 | 0.589 5 | 0.580 4 | 0.570 1 |
| Tr2-05 | 0.435 7 | 0.424 5 | 0.415 2 | 0.413 1 | 0.462 8 | 0.424 5 | 0.422 4 | 0.417 3 |
| Tr3-05 | 0.452 7 | 0.450 5 | 0.428 2 | 0.428 2 | 0.484 8 | 0.450 5 | 0.445 4 | 0.425 1 |
| Tr2-10 | 0.568 7 | 0.566 5 | 0.523 2 | 0.523 2 | 0.576 8 | 0.567 6 | 0.546 4 | 0.512 1 |
| Tr2-10 | 0.577 5 | 0.590 8 | 0.518 2 | 0.517 1 | 0.586 7 | 0.585 6 | 0.529 4 | 0.523 3 |
| avg rank | 6.12 | 5.88 | 2.00 | 1.50 | 7.88 | 5.62 | 3.88 | 2.25 |

Hamming Loss:

| Dataset | IC | CC | MEMM | VCC | RAkd | PCC | CT | SICL |
|---|---|---|---|---|---|---|---|---|
| Elec05 | 0.300 7 | 0.294 5 | 0.279 1 | 0.279 1 | 0.313 8 | 0.298 6 | 0.290 4 | 0.288 3 |
| Elec10 | 0.357 5 | 0.357 5 | 0.332 1 | 0.332 1 | 0.372 8 | 0.357 5 | 0.345 3 | 0.350 4 |
| Tr1-05 | 0.543 5 | 0.550 7 | 0.538 3 | 0.536 2 | 0.649 8 | 0.547 6 | 0.542 4 | 0.532 1 |
| Tr1-10 | 0.600 5 | 0.600 5 | 0.592 3 | 0.591 2 | 0.634 8 | 0.600 5 | 0.593 4 | 0.588 1 |
| Tr2-05 | 0.437 7 | 0.427 5 | 0.419 2 | 0.417 1 | 0.467 8 | 0.427 5 | 0.424 4 | 0.420 3 |
| Tr3-05 | 0.456 5 | 0.456 5 | 0.434 2 | 0.434 2 | 0.492 8 | 0.456 5 | 0.451 4 | 0.431 1 |
| Tr2-10 | 0.573 5 | 0.576 6 | 0.530 2 | 0.531 3 | 0.583 8 | 0.578 7 | 0.554 4 | 0.522 1 |
| Tr2-10 | 0.586 5 | 0.600 7 | 0.529 2 | 0.528 1 | 0.597 6 | 0.600 7 | 0.542 4 | 0.538 3 |
| avg rank | 5.50 | 5.62 | 2.00 | 1.62 | 7.75 | 5.75 | 3.88 | 2.12 |

Zero/One Loss:

| Dataset | IC | CC | MEMM | VCC | RAkd | PCC | CT | SICL |
|---|---|---|---|---|---|---|---|---|
| Elec05 | 0.711 8 | 0.593 5 | 0.565 1 | 0.565 1 | 0.640 7 | 0.591 3 | 0.625 6 | 0.591 3 |
| Elec10 | 0.961 8 | 0.878 4 | 0.861 2 | 0.861 2 | 0.948 7 | 0.878 4 | 0.892 6 | 0.860 1 |
| Tr1-05 | 0.910 8 | 0.892 7 | 0.867 2 | 0.869 3 | 0.875 4 | 0.889 6 | 0.875 4 | 0.859 1 |
| Tr1-10 | 0.954 8 | 0.923 5 | 0.917 3 | 0.916 2 | 0.938 7 | 0.923 5 | 0.921 4 | 0.913 1 |
| Tr2-05 | 0.749 8 | 0.667 6 | 0.631 3 | 0.630 1 | 0.642 4 | 0.666 5 | 0.672 7 | 0.630 1 |
| Tr3-05 | 0.739 8 | 0.691 5 | 0.668 2 | 0.669 3 | 0.689 4 | 0.692 6 | 0.697 7 | 0.652 1 |
| Tr2-10 | 0.934 8 | 0.840 6 | 0.800 2 | 0.800 2 | 0.895 7 | 0.839 5 | 0.835 4 | 0.781 1 |
| Tr2-10 | 0.943 8 | 0.889 6 | 0.800 1 | 0.800 1 | 0.934 7 | 0.887 5 | 0.833 3 | 0.845 4 |
| avg rank | 8.00 | 5.50 | 2.00 | 1.88 | 5.88 | 4.88 | 5.12 | 1.62 |

competitive, particularly on the simpler Electricity datasets. Both our proposed methods perform strongly also in this scenario: VCC on the Electricity data, and SICL on the Traveller data. This shows that if the complexity of more powerful base classifier can be afforded, there is no accuracy bonus in 'compensating' with a complex (e.g., fully cascaded or fully connected) structure. Rather, a simple dependency structure (such as that offered by VCC) is adequate. In other cases, SICL can make up from the lack of predictive power involved in the actual classification, by offering a number of techniques related derived from some of the successful multi-label methods.

RAkELd appears unsuited to a decision-tree classifier in this context. Probably it overfits the data. PCC performs is one of the more competitive methods from the literature, and it performs well overall, and always better than IC and CC, which supports the consensus in the literature. Nevertheless it does not outperform our proposed methods.

## 7. Conclusions

We revealed and elaborated on many connections between methods for multi-label classification and those typically used in applications involving sequential data. Our motivation was based on that drawing stronger links will be mutually beneficial for both areas of research. For example, the 'label bias' problem relating to a number of Markov models used on sequential data is in fact the same concept as 'error propagation' cited frequently in multi-label classification, with tens of models proposed to avoid it.

The insights gained from this study lead to the development of two novel approaches, which combine the benefits of several existing approaches in multi-label and sequential data, namely, linking using individual predictions in a chain-likes structure, and, and predicting labels together in a prototype-variety of approach). We tested these novel methods and found them to render promising results.

In the future we intend to investigate the effect of our proposed methods under ensemble schemes, and extend the study to consider connections between multi-output regression and parallel methods in the Markov-literature, for example Kalman and particle filters. This would address the same parallels that we have covered in this work, but for continuous output variables.

## Acknowledgements

This work was supported in part by the Aalto University AEF research programme http://energyefficiency.aalto.fi/en/

# References

[1] M.-L. Zhang, Z.-H. Zhou, A review on multi-label learning algorithms, IEEE Transactions on Knowledge and Data Engineering 26 (2014) 1819–1837.

[2] M. R. Boutell, J. Luo, X. Shen, C. M. Brown, Learning multi-label scene classification, Pattern Recognition 37 (2004) 1757–1771.

[3] J. H. Zaragoza, L. E. Sucar, E. F. Morales, C. Bielza, P. Larrañaga, Bayesian chain classifiers for multidimensional classification, in: 24th International Joint Conference on Artificial Intelligence (IJCAI '11), pp. 2192–2197.

[4] D. Kocev, C. Vens, J. Struyf, S. Deroski, Tree ensembles for predicting structured outputs, Pattern Recognition 46 (2013) 817–833.

[5] J. Read, L. Martino, D. Luengo, Efficient Monte Carlo optimization for multi-label classifier chains, International Conference on Acoustics, Speech, and Signal Processing (ICASSP) (2013) 1–5.

[6] J. Read, L. Martino, D. Luengo, Efficient Monte Carlo methods for multi-dimensional learning with classifier chains, Pattern Recognition 47 (2014) 1535–1546.

[7] D. Barber, Bayesian Reasoning and Machine Learning, Cambridge University Press, 2012.

[8] Y. Bengio, Markovian models for sequential data, Neural Computing Surveys 2 (199) 129–162.

[9] J. Hollmén, V. Tresp, Call-based fraud detection in mobile communications networks using a hierarchical regime-switching model, in: Advances in Neural Information Processing Systems 11: Proceedings of the 1998 Conference (NIPS'11), MIT Press, 1999, pp. 889–895.

[10] S. Reddy, M. Mun, J. Burke, D. Estrin, M. Hansen, M. Srivastava, Using mobile phones to determine transportation modes, ACM Trans. Sen. Netw. 6 (2010) 13:1–13:27.

[11] P. Widhalm, P. Nitsche, N. Brandie, Transport mode detection with realistic smartphone sensor data, in: Pattern Recognition (ICPR), 2012 21st International Conference on, pp. 573–576.

[12] L. Liao, D. J. Patterson, D. Fox, H. Kautz, Learning and inferring transportation routines, Artif. Intell. 171 (2007) 311–331.

[13] A. J. Viterbi, Error bounds for convolutional codes and an asymptotically optimum decoding algorithm, IEEE Transactions on Information Theory 2 (1967) 260–269.

[14] T. G. Dietterich, Machine learning for sequential data: A review, in: Proceedings of the Joint IAPR International Workshop on Structural, Syntactic, and Statistical Pattern Recognition, Springer-Verlag, London, U.K., 2002, pp. 15–30.

[15] K. Dembczyński, W. Waegeman, W. Cheng, E. Hüllermeier, On label dependence and loss minimization in multi-label classification, Mach. Learn. 88 (2012) 5–45.

[16] J. Read, B. Pfahringer, G. Holmes, E. Frank, Classifier chains for multi-label classification, Machine Learning 85 (2011) 333–359.

[17] K. Dembczyński, W. Cheng, E. Hüllermeier, Bayes optimal multilabel classification via probabilistic classifier chains, in: ICML '10: 27th International Conference on Machine Learning, Omnipress, Haifa, Israel, 2010, pp. 279–286.

[18] G. Tsoumakas, I. Katakis, I. Vlahavas, Random k-labelsets for multi-label classification, IEEE Transactions on Knowledge and Data Engineering 23 (2011) 1079–1089.

[19] A. Krogh, An introduction to hidden markov models for biological sequences, in: S. L. Salzberg, D. B. Searls, S. Kasif (Eds.), Computational Methods in Molecular Biology, Elsevier, 1998.

[20] Y. Guo, S. Gu, Multi-label classification using conditional dependency networks., in: IJCAI '11: 24th International Conference on Artificial Intelligence, IJCAI/AAAI, 2011, pp. 1300–1305.

[21] A. Kumar, S. Vembu, A. Menon, C. Elkan, Beam search algorithms for multilabel learning, Machine Learning 92 (2013) 65–89.

[22] N. Ghamrawi, A. McCallum, Collective multi-label classification, in: CIKM '05: 14th ACM international Conference on Information and Knowledge Management, ACM Press, New York, NY, USA, 2005, pp. 195–200.

[23] J. Read, L. Martino, P. M. Olmos, D. Luengo, Scalable multi-output label prediction: From classifier chains to classifier trellises, Pattern Recognition 48 (2015) 2096–2109.

[24] O. Mazhelis, I. Zliobaite, M. Pechenizkiy, Context-aware personal route recognition, in: DS '11: 14th International Conference on Discovery Science, Springer, 2011, pp. 221–235.

[25] G. Tsoumakas, I. Katakis, Multi label classification: An overview, International Journal of Data Warehousing and Mining 3 (2007) 1–13.

[26] Y. Michalevsky, A. Schulman, G. A. Veerapandian, D. Boneh, G. Nakibly, Powerspy: Location tracking using mobile device power analysis, in: J. Jung, T. Holz (Eds.), 24th USENIX Security Symposium, USENIX Security 15, Washington, D.C., USA, August 12-14, 2015., USENIX Association, 2015, pp. 785–800.

[27] K. Dembczyński, W. Waegeman, E. Hüllermeier, An analysis of chaining in multi-label classification., in: ECAI: European Conference of Artificial Intelligence, volume 242, IOS Press, 2012, pp. 294–299.

[28] J. Read, J. Hollmén, A deep interpretation of classifier chains, in: Advances in Intelligent Data Analysis XIII - 13th International Symposium, IDA 2014, pp. 251–262.

[29] S. Ben Taieb, A. Sorjamaa, G. Bontempi, Multiple-output modeling for multi-step-ahead time series forecasting, Neurocomput. 73 (2010) 1950–1957.

[30] M. Harries, Splice-2 comparative evaluation: Electricity pricing, Technical Report, The University of South Wales, 1999.

[31] N. Zorbas, D. Zissis, K. Tserpes, D. Anagnostopoulos, Predicting object trajectories from high-speed streaming data, in: 9th IEEE International Conference on Big Data Science and Engineering (IEEE BDSE-15).

[32] J. Read, P. Reutemann, B. Pfahringer, G. Holmes, MEKA: A multi-label/multi-target extension to Weka, Journal of Machine Learning Research 17 (2016) 1–5.